\documentclass[letterpaper, 10 pt, conference]{ieeeconf}
\usepackage{amsmath,amsfonts}
\usepackage{algorithmic}
\usepackage{algorithm}
\usepackage{array}
\usepackage[caption=false,font=normalsize,labelfont=sf,textfont=sf]{subfig}
\usepackage{textcomp}
\usepackage{stfloats}
\usepackage{url}
\usepackage{verbatim}
\usepackage{graphicx}
\usepackage{cite}
\usepackage[table,svgnames]{xcolor} 
\usepackage{booktabs}
\usepackage{float}
\usepackage[colorinlistoftodos, disable]{todonotes}

\usepackage[breaklinks,colorlinks,bookmarksopen,bookmarksnumbered,citecolor=red,urlcolor=red,linkcolor=red]{hyperref}
\usepackage{comment}
\IEEEoverridecommandlockouts 

\newcommand{\dakarai}[1]{\todo[inline,color=purple!40]{#1}}
\newcommand{\alexis}[1]{\todo[inline,color=pink!40]{#1}}

\newcommand{\wenzhen}[1]{\todo[inline,color=blue!40]{#1}}

\begin{document}

\title{Multimodal Socially Aware Framework for Human-Robot Interaction in Virtual Reality}

\author{Richard Xu$^{1}$, Dakarai Crowder$^{1}$, Alexis E. Block$^{2}$, and Wenzhen Yuan$^{1}$
\thanks{$^{1}$ are with University of Illinois at Urbana-Champaign, Champaign, IL, USA
        {\tt\small \{rickxu2,  dcrowd3, yuanwz\}@illinois.edu}}%
\thanks{$^{2}$ is with Case Western Reserve University, Cleveland, OH, USA
        {\tt\small \{alexis.block\}@case.edu}}%
\thanks{ This work has been submitted to the IEEE for possible publication. Copyright may be transferred without notice, after which this version may no longer be accessible. \textit{Richard Xu and Dakarai Crowder contributed equally to this work}}
}

\markboth{Journal of \LaTeX\ Class Files,~Vol.~14, No.~8, August~2021}%
{Shell \MakeLowercase{\textit{et al.}}: A Sample Article Using IEEEtran.cls for IEEE Journals}


\maketitle

\begin{abstract}
Social robots must coordinate speech and movement while interpreting multimodal human input. We present a multimodal framework that integrates speech, vision, and touch to generate contextually appropriate robot speech and movement. We evaluated the framework in virtual reality with 18 participants across five social interaction scenarios, comparing its performance against a human-teleoperated baseline. Although participants rated the autonomous system above the neutral midpoint for both naturalness and social appropriateness, the teleoperated baseline received significantly higher ratings on both measures. Behavioral coding and qualitative feedback revealed that conversational timing, clarification of ambiguous requests, and spatial coordination were important factors contributing to this performance gap. These findings highlight key challenges in developing autonomous robots capable of natural, multimodal social interaction.
\end{abstract}


\section{Introduction}
As robots become more integrated into daily life (e.g., in homes, schools, hospitals, public spaces), how can they behave in a \textit{socially appropriate} and \textit{natural} manner? Endowing robots with this ability is no simple task, as they must continuously coordinate speech, body movement, and physical contact while adapting to interruptions, misunderstandings, and shifting social contexts. Humans manage this coordination effortlessly, yet this task remains extremely challenging for robots.


Robots struggle to integrate multi-modal data streams, adapt to unpredictable human behavior, and respond through coordinated modalities, which are necessary for an interaction to feel natural. Previous researchers have attempted to solve this problem by fusing vision, eye gaze, and speech \cite{nichols2026holistic, huang2025emotion, bao2025hierarchical, chen2026real, zhang2025social}. While humans use social touch as an important modality to communicate with each other, using touch as an input in robotics remains underexplored. Furthermore, most of these systems are validated through single-turn interactions or through videos of a robot interaction that participants watch, rather than through first-person interaction with the robot \cite{wang2024lami,nichols2026holistic,bao2025hierarchical}. The absence of multi-turn interactions with an embodied robotic agent makes it difficult to understand the extent to which these systems are capable of social intelligence and to identify the system limitations contributing to poor perceptions of social intelligence.

\begin{figure}

  \centering
  \includegraphics[width=\linewidth]{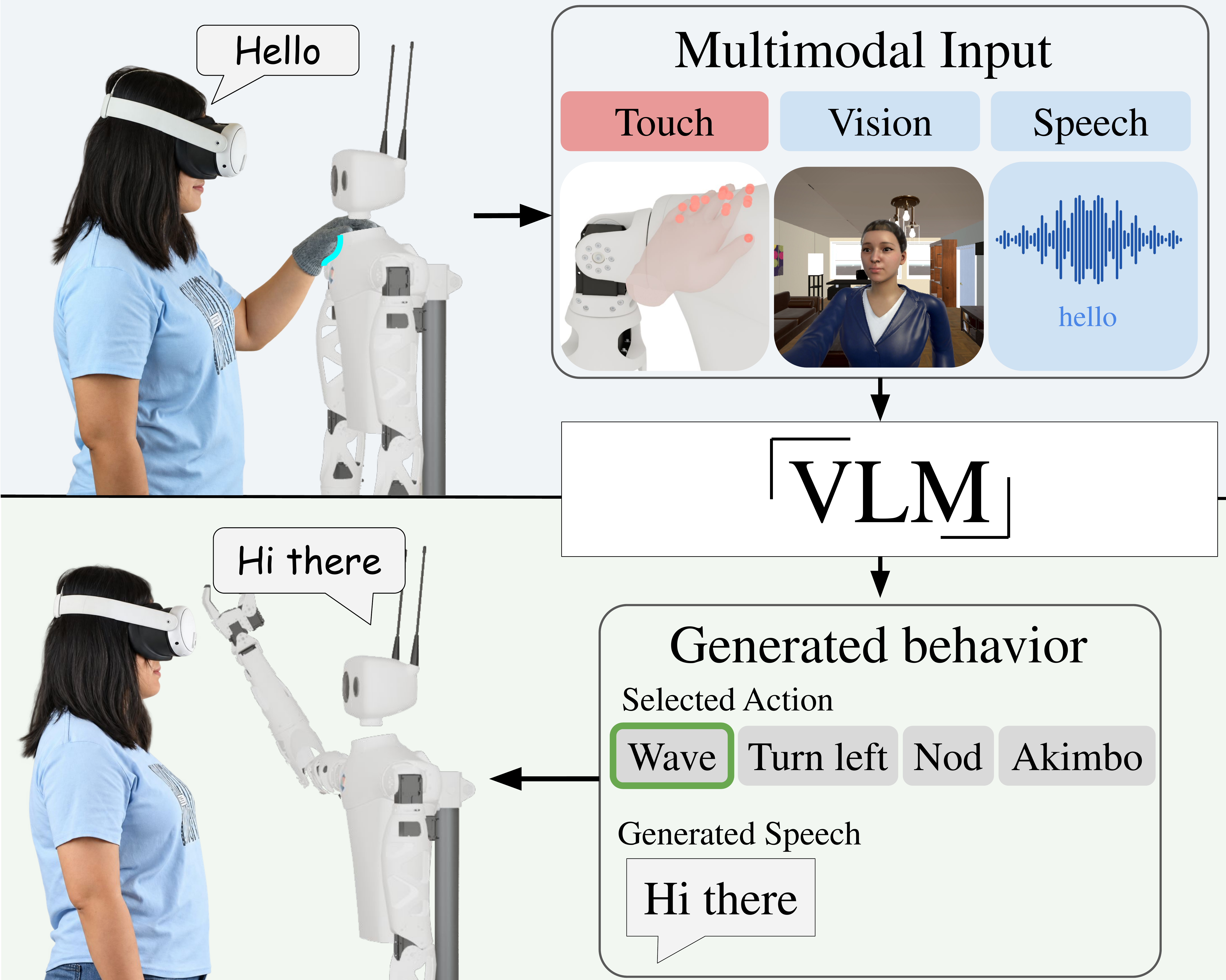}
  \vspace{-.7cm}
  \caption{We created a multimodal socially aware framework that can interact with humans. The system takes touch, vision, and speech as input and then produces socially appropriate robot behaviors.}
  \label{fig:teaser}
  \vspace{-.4cm}
\end{figure}

To enable robots to interact naturally and appropriately across multiple modalities, we present a multimodal socially aware interaction framework that integrates speech, vision, and social touch to generate coordinated speech and robot movement. We represent all modalities as text or image and keep a running interaction history to enable multi-turn interactions. We deploy this framework on a mobile humanoid robot in a simulated environment where participants interact with it through a virtual reality (VR) headset.

We evaluate the framework in VR rather than on a physical robot for five reasons: (1)~VR provides a consistent and controlled environment across scenarios and participants; (2)~VR allows instant switching between scenario contexts without physically resetting the environment; (3)~simulation provides dense contact signals over the robot's entire surface, which are difficult to obtain reliably from onboard tactile sensors; (4)~the immersive headset and haptic gloves preserve the visual and tactile experience as closely as possible; and (5)~a virtual robot removes the risk of injury during close-proximity contact.

We evaluate the system across five social interaction scenarios using a within-subject comparison between our autonomous system and a human-teleoperated baseline. For each condition, the participants rated the perceived naturalness and social appropriateness of the interaction and shared open-ended feedback regarding their expectations of robot behavior. Additionally, interaction failures in each condition from video recordings were coded. 

Our results show that participants generally perceived the autonomous system as socially appropriate, although the teleoperated baseline received significantly higher ratings for both appropriateness and naturalness. 
Behavioral analysis revealed more missed acknowledgments, repeated user attempts, interruptions, and physical touch in autonomous interactions, consistent with participants attempting to repair communication breakdowns through multiple modalities.

In this work, we contribute
\textbf{a multimodal VLM-driven framework} that integrates speech, vision, and social touch to generate socially aware robot behavior.

\section{Related Work}
\subsection{Social-Physical Human-Robot Interaction}

Prior work has explored social touch across different robotic embodiments. Zoomorphic robots usually support free-form touch interaction, but respond from a predefined repertoire of pet-like behaviors \cite{Shibata2011, guo2023, Jeong2015}. Humanoid robots supporting social touch also respond through a predefined set of reactions \cite{wood2021developing, martinez2016expressive, lehmann2022hey}. One study advances this approach by modeling a continuous emotional state that evolves with the interaction history, but the meaning assigned to each touch and the robot's expressive repertoire remain predefined \cite{burns2025creating}. However, a humanoid form may raise expectations of higher social intelligence \cite{cooney2015importance}. Other works focus on touch within structured activities, such as hugging \cite{Block2021-gp, Block2023-bw, akiyoshi2025implementation} or hand-clapping games \cite{Fitter2020-di}. Still, a review revealed that most works that combine speech and touch used Wizard-of-Oz control or scripted routines \cite{howcroft2026}. Thus, a robot should have the ability to interpret touch in the context of the interaction and respond autonomously beyond a predefined set of behaviors.

\subsection{Physically Embodied Agents for Social Interaction}
Generating socially appropriate robot behavior has traditionally relied on rule-based systems grounded in human behavioral research \cite{cassell2001beat, huang2012robot} or on models learned from human interaction data \cite{nyatsanga2023comprehensive}, but large language models (LLMs) and vision-language models (VLMs) are increasingly used as high-level reasoning engines for social robots due to their ability to reason over natural language and environmental context. These models can coordinate multimodal behaviors like speech, gestures, and head movement \cite{wang2024lami, allgeuer2024robots}, as well as physical assistance \cite{allgeuer2024robots}. Beyond action selection, LLMs have been used to generate expressive verbal and nonverbal behavior and to model social dynamics such as turn-taking, posture, and gaze \cite{nichols2026holistic, huang2025emotion, bao2025hierarchical}. Recent work shows these capabilities can be achieved using smaller, locally deployed models and refined through self-feedback mechanisms \cite{lim2026robot, galatolo2025simultaneous}. This body of work demonstrates that LLMs can generate socially aware robot behaviors; however, these systems rely almost entirely on visual and audio modalities, with social touch largely unexamined.  Evaluations of subjective ratings or isolated behaviors offer little insight into where interactions break down or how users respond when they do.

\subsection{Virtual Reality for Human-Robot Interaction}
Developing robotic systems that interact with humans can be time-consuming and costly and may introduce safety concerns during early-stage testing. Virtual reality provides an alternative platform for evaluating robot behaviors in a controlled environment and has been adopted in HRI research \cite{crowder2026requirement, haney2026perceived, albeaino2025assessing}. Prior work has shown that perceptions of robots are often preserved across live, VR, and screen-based presentations \cite{mara2021user}, and that trust and engagement measures may not significantly differ between real-world, VR, and screen-based interactions \cite{plomin2023virtual}. Although some differences have been observed, such as increased interpersonal distances in VR \cite{li2019comparing}, these findings suggest that VR is a useful and experimentally controlled platform for studying human-robot interaction.

\section{System}
\begin{figure}[h]
  \centering
  \includegraphics[width=\linewidth]{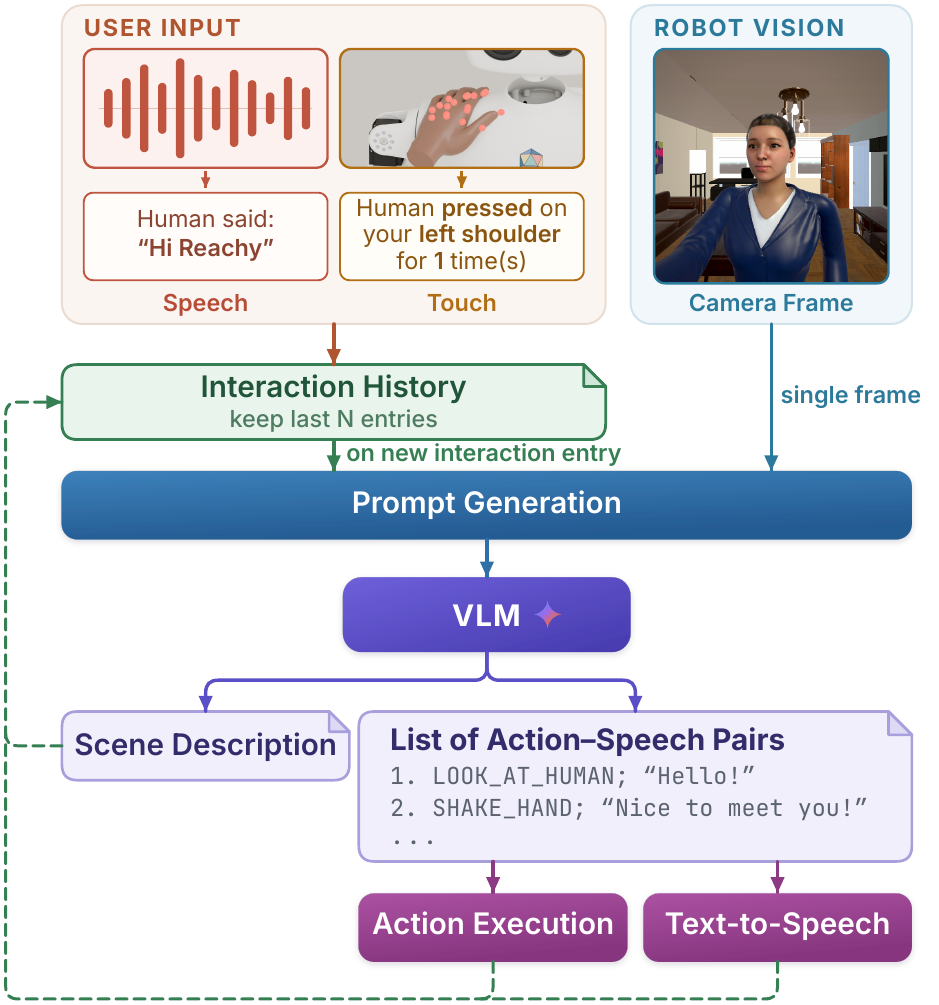}
  \vspace{-.8cm}
  \caption{System overview. Speech, touch, and vision inputs are composed into a VLM prompt; the VLM produces action–speech pairs for execution and speech, with an interaction history conditioning subsequent turns.}
  \label{fig:system}
  \vspace{-5pt}

\end{figure}

Fig.~\ref{fig:system} presents the proposed system, which enables multi-modal social interaction between a human and a robot. The system integrates speech, tactile, and visual data as inputs and uses a pre-trained VLM to output a sequence of action-speech pairs. Speech and tactile inputs act as triggers that prompt a new VLM response, whereas visual input is included in the prompt only as contextual information and does not by itself initiate a response. For example, asking the robot a question will trigger a VLM response, but waving to the robot will not. The system maintains a chronological textual history of up to N = 200 recent interaction entries, which is included in each prompt and cleared at the start of each scenario. This summary is included in the prompt to provide history to the VLM and preserve the context necessary for multi-turn interaction. The system was deployed on a Meta Quest 3 headset with Meta XR SDK hand tracking, running in a Unity simulation environment with Pollen Robotics' Reachy 2021 as the virtual robot embodiment. Speech was processed using Deepgram Nova-3 for transcription and a voice-cloning Qwen3-TTS model for speech generation, with Gemini-3-flash serving as the pre-trained VLM.

\subsection{Multi-modal Perception \& Input Processing}

The system collects speech, touch, and vision data and performs any necessary pre-processing.

For speech input, audio is collected through the VR headset microphone and transcribed to text in real time using the Deepgram API. A finalized transcription is returned via Deepgram's built-in endpointing/voice-activity detection once a brief end-of-speech pause is detected, which triggers prompt generation and VLM inference.

\begin{figure}[h]
  \centering
  \includegraphics[width=\columnwidth]{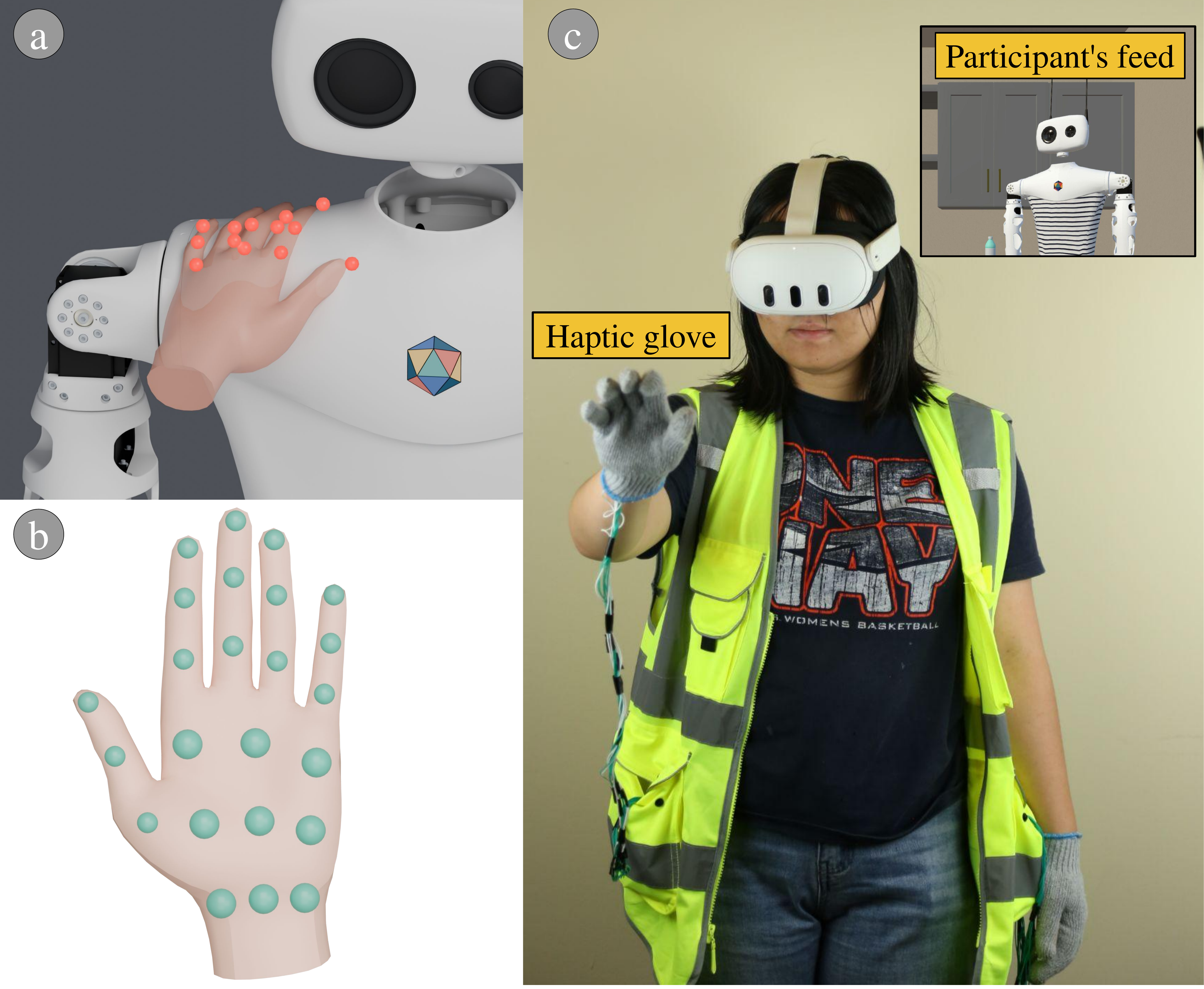}
    \vspace{-.7cm}
  \caption{Contact sensing between the tracked hand and the simulated Reachy. (a) The hand (translucent) resting on the robot's shoulder; red markers are the physics engine's contact points on the robot surface at a single frame. (b) The same hand open, palm toward the viewer, with its colliders at true scale: three per finger ($r=7$ mm) and a $3\times3$ palm grid ($r=10$ mm). (c) The participant was equipped with haptic gloves that indicated when contact occurred with the robot.}
  \label{fig:hand_colliders}

\end{figure}


To obtain social touch input in the form of natural language so it can be understood by the VLM, we first record collision events in the simulated VR environment. Then, we use a classifier to determine the gesture performed, contact location, and number of repetitions. In the simulated environment, each of the user’s hands is represented using 24 spherical colliders: three spheres of 1.4~cm in diameter along each finger and a 3×3 grid of 2.0~cm spheres on the palm, at locations shown in Fig.~\ref{fig:hand_colliders}. Each collider is attached to a skeletal point on the hand mesh through a spring-damper system. Collisions between each on-hand collider and the robot collision mesh are detected through Unity's built-in collision detection \cite{unity2021collision} and used to generate a spatiotemporal point cloud. A point $(x,y,z,F,t)$ includes three spatial coordinates $x,y,z$, the force magnitude $F$, and the timestamp $t$ of the current frame. Force is obtained by reading output from Unity's physics engine.

\begin{figure}[h]
  \centering
  \includegraphics[scale=1]{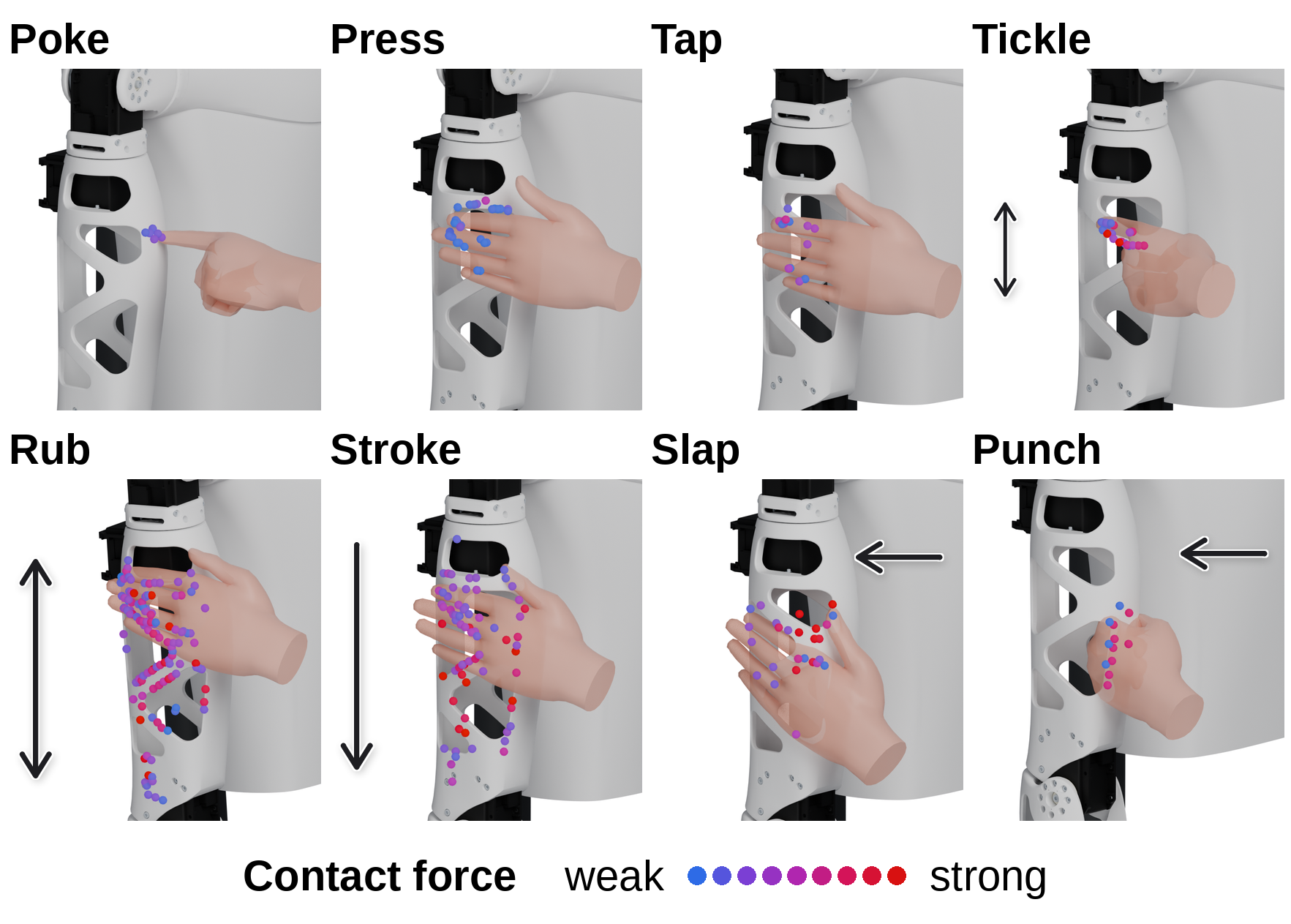}
    \vspace{-.4cm}
  \caption{The eight touch gestures, captured in VR. Translucent mesh: the tracked hand at first contact. Markers: all contact points recorded over the gesture episode, on the robot surface, colored by contact strength on a shared scale.}
  \label{fig:gestures}
  \vspace{-.1cm}
\end{figure}

When a participant makes contact with the robot, the system collects the point cloud data in each frame until it detects no contact for 0.5 seconds. The system then parses the data as a single \textit{touch event} and forwards it to the gesture classification module.

To classify the touch event, selected features are extracted from the point cloud, including duration, force, number of contacts, centroid movement, and contact location. We collected a dataset of 1756 samples across 8 social touch gestures (listed in Fig.~\ref{fig:gestures}) and 7 pilot participants before the experiment. We classify the performed gesture by averaging class probabilities from three classifiers: Gradient Boosting, an RBF-kernel SVM, and Extra Trees. Each sample in the dataset is summarized by a 60-dimensional feature vector composed of pre-defined statistical features, of which the most discriminative are the total path length of the contact centroid across frames, the event duration, the per-frame contact count, and the mean and peak force — i.e., whether the contact slides across the body, how long and how hard it presses, and over how large an area. Our leave one subject out accuracy was 78.2 ± 8.2 \%.

\subsection{Social Reasoning \& Behavior Generation}
To generate socially aware robot motion and speech, we used a VLM with chain-of-thought reasoning (Gemini-3-flash). Whenever the speech or tactile module triggers prompt generation, a new prompt is formulated. Each text prompt consists of three parts: a fixed pre-prompt, the robot's current execution state, and interaction history. In addition to the text prompt, the VLM also receives an image captured by the robot's eye camera in the virtual environment at the time of prompt generation.

\begin{enumerate}
    \item \textbf{The pre-prompt} gives relevant contextual information about the interaction and a list of motion primitives as possible output actions. It also includes an instruction to return a \textit{should\_interrupt} flag indicating whether the new input warrants overriding the robot's current behavior.
    \item \textbf{The execution state} describes the action the robot is currently performing and the motions queued to follow, giving the VLM the context needed to make that interruption decision.
    \item \textbf{The interaction history} is a chronological list of past interaction entries in text form, with the entry types listed in Table~\ref{tab:history entries}. The current speech and touch inputs are appended to this history as a transcription or a summary of the touch event, including gesture class, classifier confidence, number of repetitive touches, and the robot body part touched.
\end{enumerate}

\begin{table}
\caption{Interaction History Entries}
\label{tab:history entries}
\centering
\renewcommand{\arraystretch}{1.3}
\begin{tabular}{
  >{\raggedright\arraybackslash}p{18mm}
  >{\raggedright\arraybackslash}p{26mm}
  >{\raggedright\arraybackslash}p{34mm}}
\toprule
\textbf{Type} & \textbf{Content (text-only)} & \textbf{Added to history when} \\
\midrule
Action-Speech Pair & Motion primitive and speech content from VLM output (started/finished) & Motion primitive execution and speech output started/finished \\
Image Description  & Content of image captured by robot eye camera & VLM outputs description for image captured at time of prompt generation \\
Touch Event        & Touch gesture type with confidence, number of touches, and body part touched & Touch on the robot finishes and triggers prompt generation \\
Speech Event       & Transcription of spoken message from human & Human speech finishes and triggers prompt generation \\
\bottomrule
\end{tabular}
\vspace{-17pt}

\end{table}

To ensure the system responds quickly to any new instruction or intervention from the human user, the system handles new inputs through three separate mechanisms.
\begin{enumerate}
    \item \textbf{Request supersession:} When a new speech transcript or touch event arrives, any in-flight VLM request is invalidated and its response discarded; touch and speech arriving together are coalesced into one prompt.
    \item \textbf{VLM-decided motion interruption:} Each VLM response includes a \textit{should\_interrupt} flag. If set, the running motion is canceled, the robot resets to a safe pose, the motion queue is cleared, and the new action–speech pairs are executed; if not set while the robot is busy, the current behavior continues, and the input is retained only as context.
    \item \textbf{Speech interruption:} Robot speech is cut off immediately if the user starts speaking, or skipped entirely if the user is already speaking.
\end{enumerate}

\section{Evaluation}

To evaluate the system and assess the perceived naturalness and social appropriateness of the interactions, we conducted a VR user study using a simulated Pollen Robotics Reachy (2021). Participants completed ten counterbalanced scenarios in total: five with our proposed autonomous system and five with the teleoperated upper baseline. We selected a VR setup to enable collision detection, safe physical interaction, and controlled environmental conditions during human–robot contact, while also facilitating standardization across experimental conditions.

\subsection{Setup}

The study included a custom haptic glove system and a monitor displaying a mirrored view of the participant's VR environment for the experimenter. A camera, facing the participant at an approximate height of five feet, was placed in the room to record the participant.

The custom haptic gloves were constructed using six linear resonant actuators per glove, positioned at the fingers and palm and controlled by an Arduino Nano microcontroller communicating via Bluetooth \cite{crowder2026requirement}.
The virtual robot was represented as an articulated body to allow compliant interaction; participants could move the robot's arms and head through direct contact, though the base remained rigid unless a base movement command was being executed. Participants could also interact with the robot via speech.
To avoid the need to remove the headset between scenarios, a questionnaire interface was embedded within the VR environment, allowing participants to complete all ratings in-headset.

\subsubsection*{Human-Teleoperated Baseline}
To evaluate our system, we created a human-teleoperated baseline, which serves as a goal our system strives to reach. For this system, the operator was situated in a separate room with access to the robot's camera feed and the participant's audio, allowing real-time perception of the interaction. The operator responded using their own voice and selected movements from the same set of motion primitives available to the autonomous system through a web-based control interface. A single trained operator conducted all teleoperated sessions to ensure consistency across participants. Our autonomous system used text-to-speech (TTS) output in place of the operator's voice. To minimize bias from using different voices across conditions, we used a voice-cloning TTS model that imitates the teleoperator's voice using pre-recorded voice clips. LLM generation accounted for most of the response latency (M = 2.3 s, SD = 0.5 s; 62\% of the 3.7 s mean), followed by TTS synthesis (M = 0.85 s, 23\%).
\dakarai{this last sentence feels misplaced}
\subsection{Interaction Scenarios}

\begin{figure}[h]
  \centering
  \includegraphics[width=\linewidth]{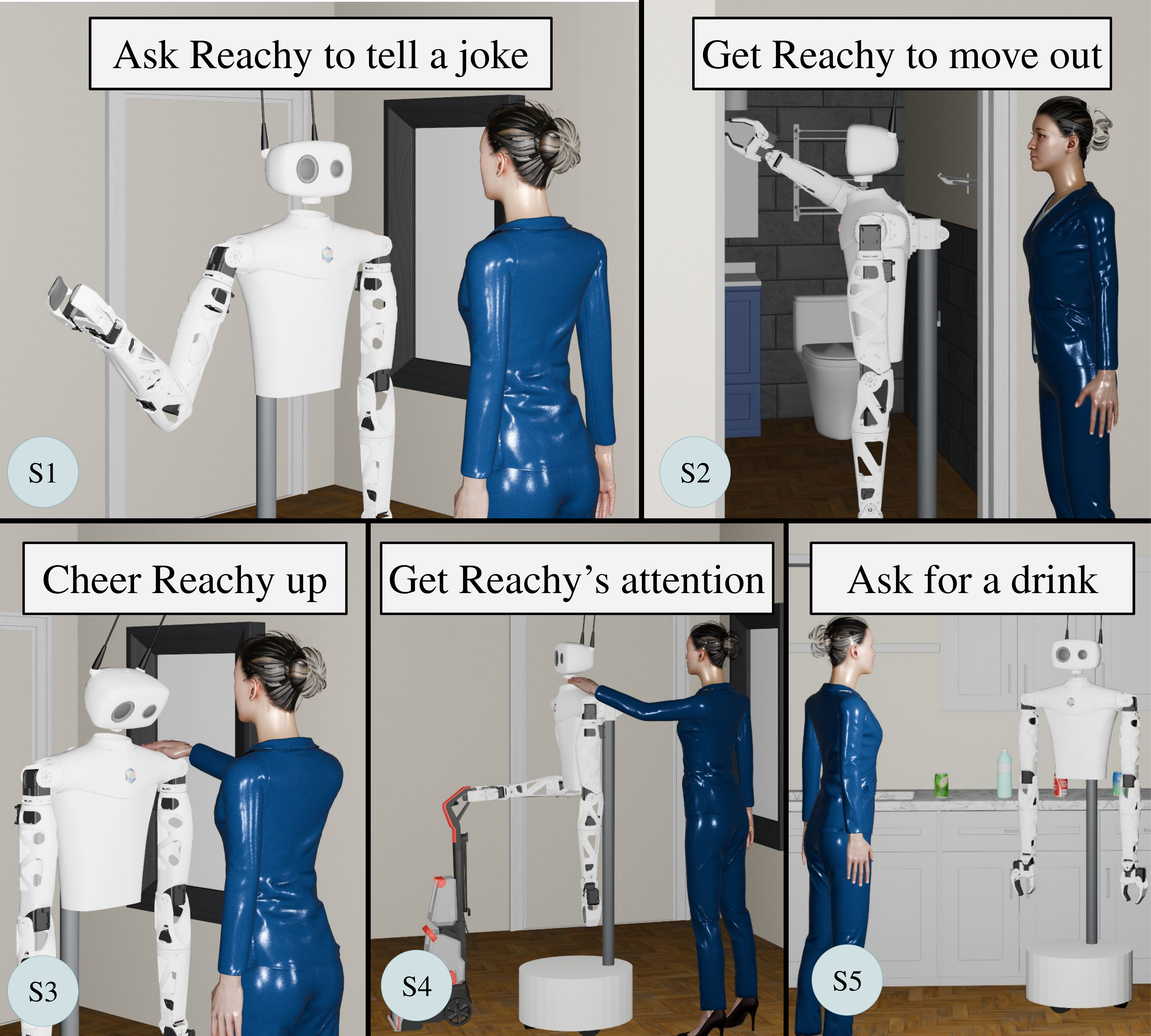}
  \vspace{-.6cm}
  \caption{Scenarios: participants completed five different scenarios. (S1) The participant is prompted to ask Reachy for a joke. (S2) Reachy is cleaning the wall but standing in front of the bathroom (the doorway on the right); the participant is prompted to get Reachy to move.  (S3) Reachy is depicted as sad, and the participant is asked to cheer the robot up. (S4) Reachy is vacuuming, and the vacuum is making a loud noise; the participant is asked to get the robot's attention. (S5) There are drinks on the counter, and the participant is prompted to ask Reachy to pass them a drink.}
  \label{fig:scenes}
\vspace{-.6cm}
\end{figure}



We created five different social scenarios that were chosen to elicit different types of social interactions. We designed scenarios based on tasks people want home robots to perform \cite{4650714}, spanning functional and emotional interactions (Fig.~\ref{fig:scenes}). Scenarios S2 and S4 encouraged physical interaction without explicitly requiring touch: in S2, the robot blocked a narrow passage while dusting, while in S4, vacuum noise discouraged verbal communication.

\subsection{User Study Procedure}

\wenzhen{Probably using bullet point for this section}
\alexis{It looks like you're missing quite a few steps in the procedure here: participant arrival, ensure meeting inclusion/exclusion criteria, informed consent, turning on video cameras, etc. that all had to come before being fitted with headset and gloves. it's super important to describe.} 
This experiment was approved by our university's Institutional Review Board. Upon arrival, participants (age 18+) were told the general purpose of the study and provided informed consent, including consent to video recording. Participants were fitted with the VR headset and haptic gloves, and then the camera was turned on. Upon entering the virtual environment, each participant’s height and arm length were calibrated to ensure accurate sizing of their virtual avatar. Participants completed a familiarization task designed to reduce the novelty effect of the VR environment. During this phase, they were allowed to freely interact with the robot. There was no time limit during this task; participants were given an example prompt to introduce themselves to the robot and were free to continue until they felt comfortable.

Following familiarization, participants completed all five interaction scenarios twice, once for the teleoperated upper baseline and once for the autonomous system. The participants were blind to the experimental condition. 
\alexis{this is a very strange way to counterbalance the presentation order. you might want to defend this choice a little more. a reviewer will ask why you didn't just counterbalance all 10 conditions instead}
A Latin square design was used to counterbalance the order of the five interaction scenarios. Within each scenario, participants completed both the autonomous and teleoperated conditions consecutively, with the order of the two conditions counterbalanced across participants. We recruited 19 participants. However, one male participant's data were excluded after they reported randomly selecting questionnaire responses; thus, our resulting dataset included 18 participants in total (M = 23.17 years, SD = 3.75 years), with 6 men (M = 22.5 years, SD = 2.59 years), 10 women (M = 23.1 years, SD = 4.61 years), and 2 non-binary participants (M = 25.5 years, SD = 0.71 years). After each scenario, the participant filled out a short questionnaire assessing their experience. After all ten scenarios, participants completed a demographics survey and GAToRS \cite{koverola2022general}. Participants received a \$15 Amazon gift card as compensation. The experiment lasted approximately one hour.

\wenzhen{I think it's helpful to show what the interface of the questionnaire part is like. You can probably show it in the video or in the supplementary materials.}

\subsection{Measures and Analyses}
After each interaction, participants rated the perceived naturalness and appropriateness of the interaction using continuous slider scales. 
\begin{itemize}
    \item The robot behavior seemed: Unnatural – Natural
    \item The robot behavior seemed: Inappropriate – Appropriate 
\end{itemize}Participants were given the opportunity to provide open-ended feedback describing how they would have preferred the robot to respond during the interaction after each scenario, a total of ten times. One author reviewed all open-ended responses, inductively generated codes, and grouped related codes into categories.

To explore behavioral differences that might explain participants' differing ratings of the two systems, behavioral coding of video recordings was performed. One author developed a codebook and coded all recordings. Codebook development occurred in two phases. First, the author watched all recordings from both conditions once for familiarization, then again to inductively identify interaction breakdowns and indicators of breakdowns. Throughout this work, we operationalized interaction breakdowns as observable failures in communication or task execution, or behaviors suggesting such a failure, such as a participant repeating themselves. In the second phase, the author reviewed all open-ended participant responses to identify additional behaviors that participants perceived positively or negatively. Candidate codes from this phase were retained if they could be unambiguously observed from the videos, which the author confirmed by reviewing the recordings again. The full set of behavioral coding categories is reported in Figure~\ref{fig:natural}(C). The author then used the finalized codebook to code all recordings.

Interaction breakdowns were analyzed using a negative binomial regression to model the total number of interaction breakdowns with condition, scenario, and their interaction as predictors, and standard errors clustered by participant to account for repeated measures. A Wilcoxon signed-rank test was used to compare participants' mean total-error counts between conditions.

Slider ratings for naturalness and appropriateness were analyzed using linear mixed-effects (LME) models with condition, scenario, and their interaction as fixed effects, participant as a random intercept, and prior robot attitude as a covariate. To test whether our autonomous system's ratings differed from the neutral midpoint (0 on the continuous scale from -1 to +1), we fit another LME model with a fixed intercept and a random intercept for participant.

\section{Results}
\wenzhen{A general problem about the result section: don't only list the numbers, but show your conclusion, explain what information you want to prove through the numbers. }
\alexis{Perhaps this is a stylistic choice or a consequence of the venue you're targeting, but I've always been told to keep results and discussion completely separate. Present the results unbiased and without interpretation to allow the reader to draw their own conclusions about what the numbers mean. Then in the discussion is your opportunity to interpret them for the reader and tell them what you think it means and try to convince/prove it to them. Because HRI requires human participants, and looks that their subjective evaluations, it's especially important to keep your interpretation/evaluation out of the results, perhaps more so than in typical robotics}

\subsection{Perceived Interaction Quality}

\begin{figure*}[h]
  \centering
  \includegraphics[width=\linewidth]{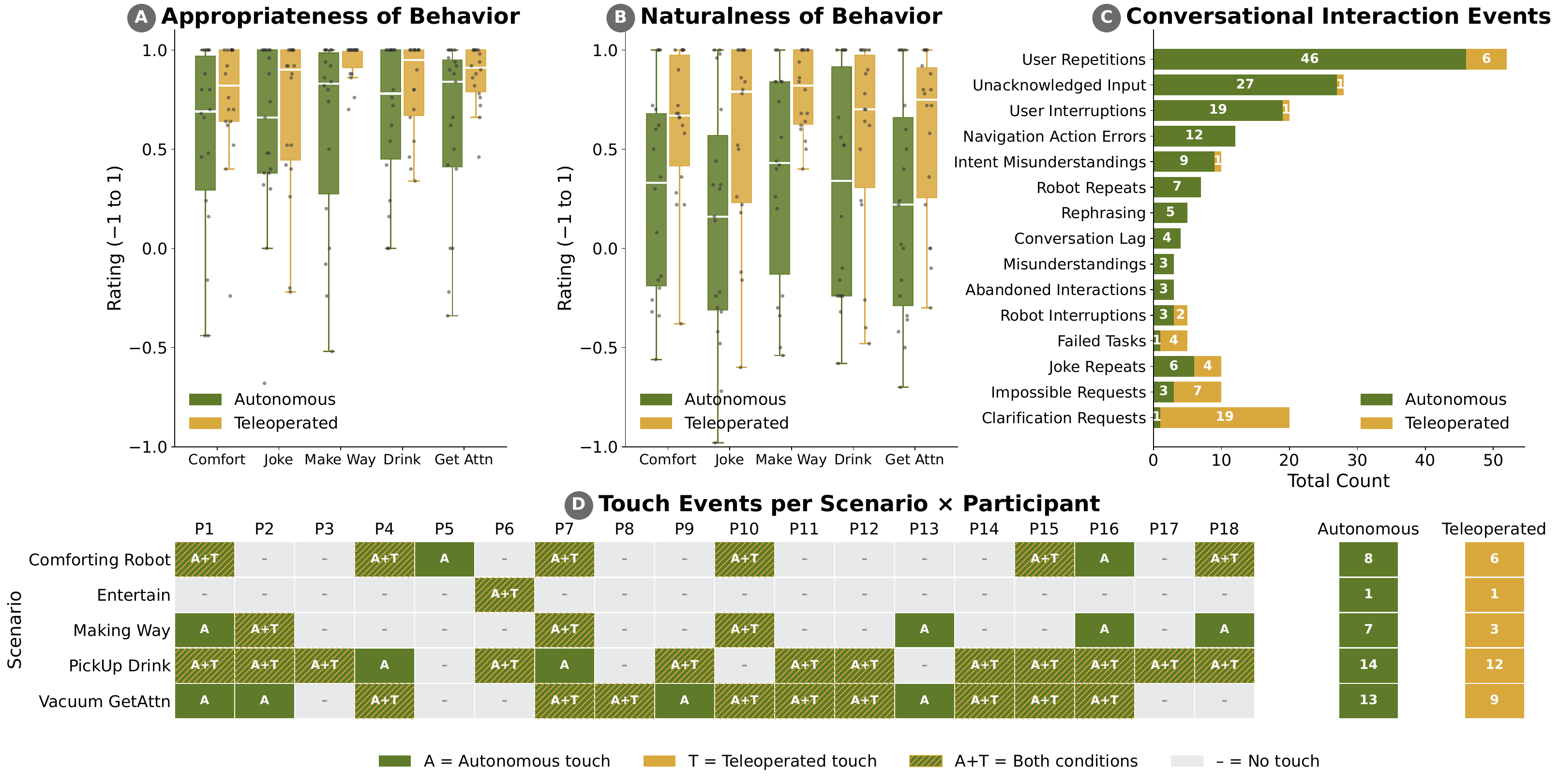}
  \vspace{-.8cm}
  \caption{Questionnaire and behavioral coding results. Per-scene ratings of (A) the appropriateness and (B) the naturalness of the autonomous system and the teleoperated baseline. (C) Interaction events coded for both the participant and the robot. The first 12 behaviors are considered interaction breakdowns, whereas the final three are considered positive interaction behaviors. (D) Whether a participant touched the robot under a condition for each scenario. There were no instances in which only the teleoperated baseline was touched. \wenzhen{You should add (a)(b)(c) etc. to the sub figures; text too small and hard to read, part of the reason is that the color is not obvious and vibrant; somewhere you should explain the scores are in the range of 0-1.}\wenzhen{The third plot is hard to understand. You need to explain what it shows}}
  \dakarai{how about now}
  \wenzhen{Caption still a little jargonized. Try to make things clearer}
  \label{fig:natural}
  \vspace{-.4cm}
\end{figure*}
\wenzhen{This subsection description is somehow confusing. You should explain at highlevel what it shows -- fundamentally you want to prove that the methods work well, and explain how the numbers support the conclusion}
\wenzhen{You should also note the label setting. You refer the proposed method as "autonomous", which might be confusing for people who jump many texts and goes to the figures directly. So a better name could be <Name of the paper's method> or "ours", and label teleoperated as "upper baseline" or some name}
\wenzhen{this is as expected, since you should consider the teleoperation data is a "ceiling score", a reference of how the best your system can reach. You should explain it. }

\begin{table}[H]
\centering
\caption{Rating Means and LME Results}
\label{tab:lme_results}
\small

\begin{tabular}{lcc}
\toprule
& \multicolumn{2}{c}{Mean (SD)}\\
\cmidrule(lr){2-3}
 & Autonomous & Teleoperated\\
\midrule
Naturalness (N) &
0.253 (0.554) &
0.620 (0.413)\\
Appropriateness (A) &
0.605 (0.443) &
0.812 (0.276)\\
\bottomrule
\end{tabular}

\vspace{0.75em}

\begin{tabular}{lcccccc}

 & $\beta$ & SE & $z$ & $p$ & 95\% CI & $d$\\
\midrule
N &
0.370 &
0.062 &
5.94 &
$<.001$ &
[0.25, 0.49] &
0.77\\

A &
0.209 &
0.050 &
4.20 &
$<.001$ &
[0.11, 0.31] &
0.58\\
\bottomrule
\end{tabular}
  \vspace{-.55cm}
\end{table}

\begin{table}[H]
    \centering
    \caption{Autonomous Ratings vs. Scale Midpoint}
    \label{tab:ratings}
    \begin{tabular}{lccc}
        \toprule
        \textbf{Dimension} & \textbf{Mean} & \textbf{95\% CI} & \textit{p} \\
        \midrule
        Naturalness     & $+0.25$ & $[+0.08,\,+0.42]$ & $.004$ \\
        Appropriateness & $+0.61$ & $[+0.48,\,+0.73]$ & $<.001$ \\
        \bottomrule
    \end{tabular}
\end{table}



Ratings of the autonomous system were above the scale midpoint for both measures (Table~\ref{tab:ratings}), with a larger margin for appropriateness. Teleoperated interactions, however, received significantly higher ratings on both measures (Table~\ref{tab:lme_results}), with no significant condition $\times$ scenario interactions (all $p$ values $> 0.38$). 

\subsection{Where Interactions Break Down}

\wenzhen{After reading the title and several starting sentences of this subsection, I don't know what you mean and what you want to convey in this subsection. You should improve the writing to change this. }
\dakarai{how is this?}
\wenzhen{Better. Still need improvements. What exactly do you mean by "interaction" here? Why does it matter? What is "break down"? Why does it matter? You should explain them clearly in the beginning.}
To understand where our system fell short of the baseline, we coded observable interaction behaviors across both conditions and compared their frequency and composition. An interaction is any event that occurred between the human and the robot.
\alexis{Missing a lot of details about the coding process. Who is "we"? one person or two? Please see my comment about the IRR in parentheses, which if it is inter-rater reliability, should not be reported there. You must describe the coding (in the measures and analyses section) how did you develop the codebook? what percent overlap of coding did you use? how did you do training and validation to ensure both coders (if there were two) were on the same page? If only one coder was used, you must state that explicitly }
The autonomous condition had significantly more breakdowns across 18 participants ($M = 1.54$, $SD = 1.42$) than the teleoperated baseline ($M =0.17$, $SD = 0.21$, $W = 1.0$, $p<0.001$, $r = 0.99$). A negative binomial model showed the difference was largest in the vacuum scenario ($Incidence Rate Ratio = 0.11$, $p < 0.001$). The most frequent forms of breakdown (Fig.~\ref{fig:natural}(C)) were all failures of conversational timing and responsiveness. Categories pointing to incorrect behavior selection, such as intent misunderstanding and failed tasks, were uncommon, with the exception of navigation errors. In contrast, the teleoperated baseline had fewer interaction breakdowns and commonly exhibited a positive interaction behavior: the teleoperator asked for clarification when given ambiguous input.

\subsection{Touch Incidents}


Participants touched the robot in 47.8\% of the autonomous interactions and 34.4\% of teleoperated interactions. In 12 cases, participants touched the robot only in the autonomous condition; the reverse never occurred (Fig.~\ref{fig:natural}(D)). Touch served different functions across scenarios. Participants used taps, pokes, and presses to get the robot's attention in the vacuum and make-way scenarios and strokes and rubs to convey affection when comforting the robot. Incidental contact occurred when participants took the drink from the robot's gripper.

\section{Qualitative Analysis}
\alexis{this section does not describe thematic analysis at all. This appears to be more of qualitative analysis. Thematic analysis is a very specific method outlined in the braun and clarke paper, which does not appear to be followed here. You should either A. Rename this section "qualitative analysis" and remove any mention of "themes" or B. actually follow and conduct a proper thematic analysis. With either method you perform, you should additionally include some quantitization of the qualitative responses. It is insufficient to say "participants said x" you must state how many participants said something like that. For example, "Participants described verbal interaction through two conversational behaviors: timing (X out of Y participants, ZZ~\%) and clarification (X out of Y participants, ZZ~\%)." and so on. }
Beyond the observable behaviors and subjective measures, we wanted to identify recurring themes in participants' qualitative feedback that contextualized the quantitative differences between conditions. Thus, we conducted a qualitative analysis of the open-ended responses the users gave after each interaction.
\wenzhen{Explain in plain language that what you are trying to do and what you want to prove in this section}

\subsection{Verbal Interaction}
\label{sec:verbal_interaction}
Participants described verbal interaction through two conversational behaviors: timing (10 out of 18 participants) and clarification (8 out of 18 participants). Timing was discussed most often, encompassing both response latency and conversational pacing. P12 noted that, compared with the teleoperated robot, the autonomous system's responses ``were a little bit delayed [...] it would [feel] more natural [...] if the robot responded [faster].'' Pacing mattered too: during the joke scenario, participants said the teleoperated robot's pause between question and punchline felt natural, while the autonomous system's immediate delivery broke the flow, since ``the pause [was] essential to the joke'' (P5).

Clarification was another aspect of verbal interaction: 8 out of 18 participants described seeking clarification as the preferable way to handle ambiguous requests. During the drink scenario, several remarked that they would have preferred the autonomous system to ask which drink they wanted rather than choosing automatically. For instance, when P15 asked the robot to pass them a drink, they appreciated that the teleoperated robot followed up: ``I liked how this time he asked me which drink I would like." This suggests participants expected robots to resolve ambiguity collaboratively rather than independently.


\subsection{Embodied Coordination}
Fourteen out of 18 participants evaluated the robot not only by what it said but also by how it coordinated its movement and orientation within the shared environment. Expectations included moving in contextually appropriate directions, orienting toward the participant while speaking, and responding appropriately to physical touch. As one participant noted, ``it agreed to move out of the way but it continued facing the wall while talking which was a little weird as people normally turn around to face each other.'' Participants frequently described failures in movement, orientation, or touch responsiveness as disrupting the flow of interaction, indicating that appropriate embodiment formed an important component of perceived naturalness.


\subsection{Emotional Expectations}
The Comforting Robot scenario revealed differing expectations regarding the emotional role of robots. Many participants (6 out of 18 participants) found the robot's expression of sadness surprising, awkward, or unsettling, while others responded positively when the interaction felt meaningful and reciprocal. As one participant explained, ``in [their] mind, a robot shouldn't have to have human emotions, especially sad human emotions. I feel like if anything it should be happy-go-lucky all the time.''  Three out of 18 participants also described feeling uncomfortable comforting the robot, and one participant attributed the interaction to the uncanny valley. In contrast, a smaller group (3 out of 18 participants) preferred interactions in which the robot expressed emotion with sufficient context and responded reciprocally, suggesting that emotional interactions were generally accepted when they aligned with participants' expectations of the robot's social role.

\alexis{I don't see a proper discussion or limitations section for this paper. Both are important. Limitations is usually a subsection of the Discussion, never the conclusion, you want to end on a high note} 

\section{Discussion}

Participants generally perceived the autonomous system as natural and appropriate, with the mean ratings on both measures above the scale midpoint (Table~\ref{tab:ratings}). Thus, our system produced reasonable behavior across the scenarios. However, the teleoperated baseline was rated significantly higher on both measures, particularly naturalness (Table~\ref{tab:lme_results}). We found no evidence that the condition difference varied across scenarios, as the condition $\times$ scenario interactions were not statistically significant.

The behavioral coding and participants' open-ended responses suggest why this gap exists between the systems. The most common autonomous breakdowns (user repetitions, unacknowledged input, and user interruptions) reflect slow or poorly timed responses (Fig.~\ref{fig:natural}(C)), and participants even described the autonomous responses as delayed or poorly paced (Section~\ref{sec:verbal_interaction}). The system's mean response latency was 3.7~s, longer than the typical 250~ms gaps between turns in human conversation \cite{stivers2009universals}, suggesting that the delay was experienced as a communication breakdown. In contrast, interactions with the teleoperator involved few user repetitions, and the teleoperator asked for clarification when requests were ambiguous. Participants stated that they valued this behavior. Navigation errors also occurred with our autonomous system but were absent from the baseline. However, the teleoperator's familiarity with the environment may explain this gap. These findings suggest that one of the main reasons why the system has a lower rating is the pacing of the interaction. Taken together, these findings suggest that improving pacing, ensuring the system seeks clarification in ambiguous situations, and improving spatial awareness could help close this gap.

Participants touched the robot more often in the autonomous condition and never touched the robot only in the teleoperated condition (Fig.~\ref{fig:natural}(D)). The greater frequency of touch in the autonomous condition, alongside more unacknowledged input, suggests that participants may have used touch as an alternative communication channel when speech was unsuccessful. This observation motivates incorporating touch into multimodal interaction systems, although further evaluation is needed to determine whether touch improves recovery from communication breakdowns.

Interestingly, emotional expression did not converge on a single preferred behavior. Whether the robot's sadness seemed natural or unsettling seems to depend on whether the participants viewed the robot as a tool or a social agent. In this case, we cannot simply match the teleoperated baseline; we must align the behavior with each user's expectations of the robot's role.

\subsection{Limitations}
There were several limitations of this work. The teleoperator's familiarity with the virtual environment may have provided an advantage in spatial reasoning and localization. In addition, the voice-cloned TTS system had less expressive prosody than the teleoperator's natural speech, which may have influenced perceptions of naturalness. Finally, the study was conducted in virtual reality, and the findings should be validated through future studies involving physical robots in real-world environments.

\section{Conclusion}
We presented a multimodal framework integrating speech, vision, and social touch to generate socially aware robot behavior. In a VR study, participants rated the system as generally appropriate and natural and used touch alongside speech to communicate with it. The observed breakdowns and participant feedback suggest that conversational timing, clarification, and spatial coordination contribute to the gap between the autonomous system and the teleoperated baseline. Future work will address these areas and deploy the framework on a physical robot.
\vspace{-.3cm}



{\small
\bibliographystyle{IEEEtran}
\bibliography{bib}

@inproceedings{wang2024lami,
  title={Lami: Large language models for multi-modal human-robot interaction},
  author={Wang, Chao and Hasler, Stephan and Tanneberg, Daniel and Ocker, Felix and Joublin, Frank and Ceravola, Antonello and Deigmoeller, Joerg and Gienger, Michael},
  booktitle={Extended Abstracts of the CHI Conference on Human Factors in Computing Systems},
  pages={1--10},
  year={2024}
}

@inproceedings{allgeuer2024robots,
  title={When robots get chatty: Grounding multimodal human-robot conversation and collaboration},
  author={Allgeuer, Philipp and Ali, Hassan and Wermter, Stefan},
  booktitle={International Conference on Artificial Neural Networks},
  pages={306--321},
  year={2024},
  organization={Springer}
}

@inproceedings{nichols2026holistic,
  title={Holistic LLM-Based Expressive Behavior Generation for Robots},
  author={Nichols, Eric and Mejia Tobar, Miguel and Gomez, Randy},
  booktitle={Companion Proceedings of the 21st ACM/IEEE International Conference on Human-Robot Interaction},
  pages={809--813},
  year={2026}
}

@article{huang2025emotion,
  title={Emotion: Expressive motion sequence generation for humanoid robots with in-context learning},
  author={Huang, Peide and Hu, Yuhan and Nechyporenko, Nataliya and Kim, Daehwa and Talbott, Walter and Zhang, Jian},
  journal={IEEE Robotics and Automation Letters},
  year={2025},
  publisher={IEEE}
}

@article{bao2025hierarchical,
  title={Hierarchical Intention-Aware Expressive Motion Generation for Humanoid Robots},
  author={Bao, Lingfan and Pan, Yan and Peng, Tianhu and Kanoulas, Dimitrios and Zhou, Chengxu},
  journal={arXiv preprint arXiv:2506.01563},
  year={2025}
}

@article{chen2026real,
  title={Real-Time Synchronized Interaction Framework for Emotion-Aware Humanoid Robots},
  author={Chen, Yanrong and Bian, Xihan},
  journal={arXiv preprint arXiv:2601.17287},
  year={2026}
}

@inproceedings{zhang2025social,
  title={Social Agent: Mastering Dyadic Nonverbal Behavior Generation via Conversational LLM Agents},
  author={Zhang, Zeyi and Zhou, Yanju and Yao, Heyuan and Ao, Tenglong and Zhan, Xiaohang and Liu, Libin},
  booktitle={Proceedings of the SIGGRAPH Asia 2025 Conference Papers},
  pages={1--12},
  year={2025}
}

@article{lim2026robot,
  title={The Robot's Inner Critic: Self-Refinement of Social Behaviors through VLM-based Replanning},
  author={Lim, Jiyu and Yoon, Youngwoo and Park, Kwanghyun},
  journal={arXiv preprint arXiv:2603.20164},
  year={2026}
}

@article{galatolo2025simultaneous,
  title={Simultaneous text and gesture generation for social robots with small language models},
  author={Galatolo, Alessio and Winkle, Katie},
  journal={Frontiers in Robotics and AI},
  volume={12},
  pages={1581024},
  year={2025},
  publisher={Frontiers Media SA}
}

@misc{unity2021collision,
  author       = {{Unity Technologies}},
  title        = {Introduction to collision},
  howpublished = {Unity Manual, Version 2021.3},
  year         = {2025},
  note         = {Accessed: 2026-06-16},
  url          = {https://docs.unity3d.com/2021.3/Documentation/Manual/CollidersOverview.html}
}

@article{haney2026perceived,
  title={Perceived safety during human-robot interaction with an autonomous mobile robot},
  author={Haney, Justin M and Ammons, Douglas and Choi, HeeSun},
  journal={Applied ergonomics},
  volume={135},
  pages={104747},
  year={2026},
  publisher={Elsevier}
}

@article{albeaino2025assessing,
  title={Assessing proxemics impact on Human-Robot collaboration safety in construction: A virtual reality study with four-legged robots},
  author={Albeaino, Gilles and Jeelani, Idris and Gheisari, Masoud and Issa, Raja RA},
  journal={Safety Science},
  volume={181},
  pages={106682},
  year={2025},
  publisher={Elsevier}
}

@article{mara2021user,
  title={User responses to a humanoid robot observed in real life, virtual reality, 3D and 2D},
  author={Mara, Martina and Stein, Jan-Philipp and Latoschik, Marc Erich and Lugrin, Birgit and Schreiner, Constanze and Hostettler, Rafael and Appel, Markus},
  journal={Frontiers in psychology},
  volume={12},
  pages={633178},
  year={2021},
  publisher={Frontiers Media SA}
}

@article{plomin2023virtual,
  title={Virtual reality check: a comparison of virtual reality, screen-based, and real world settings as research methods for HRI},
  author={Plomin, Jana and Schweidler, Paul and Oehme, Astrid},
  journal={Frontiers in Robotics and AI},
  volume={10},
  pages={1156715},
  year={2023},
  publisher={Frontiers Media SA}
}

@inproceedings{li2019comparing,
  title={Comparing human-robot proxemics between virtual reality and the real world},
  author={Li, Rui and van Almkerk, Marc and van Waveren, Sanne and Carter, Elizabeth and Leite, Iolanda},
  booktitle={2019 14th ACM/IEEE international conference on human-robot interaction (HRI)},
  pages={431--439},
  year={2019},
  organization={IEEE}
}

@ARTICLE{Shibata2011,
  title     = "Robot therapy: a new approach for mental healthcare of the
               elderly - a mini-review",
  author    = "Shibata, Takanori and Wada, Kazuyoshi",
  journal   = "Gerontology",
  publisher = "S. Karger AG",
  volume    =  57,
  number    =  4,
  pages     = "378--386",
  year      =  2011,
  copyright = "https://creativecommons.org/licenses/by-nc/3.0/",
  language  = "en"
}

@article{guo2023,
author = {Guo, Shihui and Zhan, Lishuang and Cao, Yancheng and Zheng, Chen and Zhou, Guyue and Gong, Jiangtao},
title = {Touch-and-Heal: Data-driven Affective Computing in Tactile Interaction with Robotic Dog},
year = {2023},
issue_date = {June 2023},
publisher = {Association for Computing Machinery},
address = {New York, NY, USA},
volume = {7},
number = {2},
url = {https://doi.org/10.1145/3596258},
doi = {10.1145/3596258},
journal = {Proc. ACM Interact. Mob. Wearable Ubiquitous Technol.},
month = jun,
articleno = {56},
numpages = {33}
}

@INPROCEEDINGS{Jeong2015,
  title           = "Designing a socially assistive robot for pediatric care",
  booktitle       = "Proceedings of the 14th International Conference on
                     Interaction Design and Children",
  author          = "Jeong, Sooyeon and Santos, Kristopher Dos and Graca,
                     Suzanne and O'Connell, Brianna and Anderson, Laurel and
                     Stenquist, Nicole and Fitzpatrick, Katie and Goodenough,
                     Honey and Logan, Deirdre and Weinstock, Peter and
                     Breazeal, Cynthia",
  publisher       = "ACM",
  month           =  jun,
  year            =  2015,
  address         = "New York, NY, USA",
  copyright       = "http://www.acm.org/publications/policies/copyright\_policy\#Background",
  conference      = "IDC '15: Interaction Design and Children",
  location        = "Boston Massachusetts"
}

@INPROCEEDINGS{Block2021-gp,
  title      = "The six hug commandments",
  booktitle  = "Proceedings of the 2021 {ACM/IEEE} International Conference on
                {Human-Robot} Interaction",
  author     = "Block, Alexis E and Christen, Sammy and Gassert, Roger and
                Hilliges, Otmar and Kuchenbecker, Katherine J",
  publisher  = "ACM",
  month      =  mar,
  year       =  2021,
  address    = "New York, NY, USA",
  conference = "HRI '21: ACM/IEEE International Conference on Human-Robot
                Interaction",
  location   = "Boulder CO USA"
}

@ARTICLE{Block2023-bw,
  title     = "In the arms of a robot: Designing autonomous hugging robots with
               intra-hug gestures",
  author    = "Block, Alexis E and Seifi, Hasti and Hilliges, Otmar and
               Gassert, Roger and Kuchenbecker, Katherine J",
  journal   = "ACM Trans. Hum. Robot Interact.",
  publisher = "Association for Computing Machinery (ACM)",
  volume    =  12,
  number    =  2,
  pages     = "1--49",
  month     =  jun,
  year      =  2023,
  language  = "en"
}

@ARTICLE{Fitter2020-di,
  title={How does it feel to clap hands with a robot?},
  author={Fitter, Naomi T and Kuchenbecker, Katherine J},
  journal={International Journal of Social Robotics},
  volume={12},
  number={1},
  pages={113--127},
  year={2020},
  publisher={Springer}
}

@misc{crowder2026requirement,
      title={Requirement-Driven Design of Whole-Body Social Tactile Sensing via Virtual Human-Robot Interaction}, 
      author={Dakarai Crowder and Ruohan Zhang and Alexis E. Block and Wenzhen Yuan},
      year={2026},
      eprint={2607.11690},
      archivePrefix={arXiv},
      primaryClass={cs.RO},
      url={https://arxiv.org/abs/2607.11690}, 
}

@article{wood2021developing,
  title={Developing kaspar: a humanoid robot for children with autism},
  author={Wood, Luke J and Zaraki, Abolfazl and Robins, Ben and Dautenhahn, Kerstin},
  journal={International Journal of Social Robotics},
  volume={13},
  number={3},
  pages={491--508},
  year={2021},
  publisher={Springer}
}

@inproceedings{martinez2016expressive,
  title={Expressive touch: Control of robot emotional expression by touch},
  author={Martinez-Hernandez, Uriel and Prescott, Tony J},
  booktitle={2016 25th IEEE International Symposium on Robot and Human Interactive Communication (RO-MAN)},
  pages={974--979},
  year={2016},
  organization={IEEE}
}

@inproceedings{lehmann2022hey,
  title={Hey, robot! An investigation of getting robot’s attention through touch},
  author={Lehmann, Hagen and Rojik, Adam and Friebe, Kassandra and Hoffmann, Matej},
  booktitle={International Conference on Social Robotics},
  pages={388--401},
  year={2022},
  organization={Springer}
}

@article{burns2025creating,
  title={Creating an Affective Robot That Feels Both Touch and Emotion},
  author={Burns, Rachael Bevill and Richardson, Benjamin A and Klingenberg, Jack and Kuchenbecker, Katherine J},
  journal={IEEE Transactions on Affective Computing},
  year={2025},
  publisher={IEEE}
}

@article{cooney2015importance,
  title={Importance of touch for conveying affection in a multimodal interaction with a small humanoid robot},
  author={Cooney, Martin D and Nishio, Shuichi and Ishiguro, Hiroshi},
  journal={International journal of humanoid robotics},
  volume={12},
  number={01},
  pages={1550002},
  year={2015},
  publisher={World Scientific}
}

@article{akiyoshi2025implementation,
  title={Implementation and Evaluation of Intra-Hug Gestures During Dialogue for a Huggable Robot},
  author={Akiyoshi, Takuto and Sumioka, Hidenobu and Nakanishi, Junya and Kato, Hirokazu and Shiomi, Masahiro},
  journal={IEEE Robotics and Automation Letters},
  volume={11},
  number={1},
  pages={850--857},
  year={2026},
  publisher={IEEE}
}

@inproceedings{cassell2001beat,
  title={Beat: the behavior expression animation toolkit},
  author={Cassell, Justine and Vilhj{\'a}lmsson, Hannes H{\"o}gni and Bickmore, Timothy},
  booktitle={Proceedings of the 28th annual conference on Computer graphics and interactive techniques},
  pages={477--486},
  year={2001}
}

@inproceedings{huang2012robot,
  title={Robot behavior toolkit: generating effective social behaviors for robots},
  author={Huang, Chien-Ming and Mutlu, Bilge},
  booktitle={Proceedings of the seventh annual ACM/IEEE international conference on Human-Robot Interaction},
  pages={25--32},
  year={2012}
}

@inproceedings{nyatsanga2023comprehensive,
  title={A comprehensive review of data-driven co-speech gesture generation},
  author={Nyatsanga, Simbarashe and Kucherenko, Taras and Ahuja, Chaitanya and Henter, Gustav Eje and Neff, Michael},
  booktitle={Computer Graphics Forum},
  volume={42},
  number={2},
  pages={569--596},
  year={2023},
  organization={Wiley Online Library}
}

@INPROCEEDINGS{4650714,
  author={Ray, Celine and Mondada, Francesco and Siegwart, Roland},
  booktitle={2008 IEEE/RSJ International Conference on Intelligent Robots and Systems}, 
  title={What do people expect from robots?}, 
  year={2008},
  volume={},
  number={},
  pages={3816-3821},
  doi={10.1109/IROS.2008.4650714}}

@article{stivers2009universals,
  title={Universals and cultural variation in turn-taking in conversation},
  author={Stivers, Tanya and Enfield, Nicholas J and Brown, Penelope and Englert, Christina and Hayashi, Makoto and Heinemann, Trine and Hoymann, Gertie and Rossano, Federico and De Ruiter, Jan Peter and Yoon, Kyung-Eun and others},
  journal={Proceedings of the National Academy of Sciences},
  volume={106},
  number={26},
  pages={10587--10592},
  year={2009},
  publisher={National Academy of Sciences}
}

@article{koverola2022general,
  title={General attitudes towards robots scale (GAToRS): A new instrument for social surveys},
  author={Koverola, Mika and Kunnari, Anton and Sundvall, Jukka and Laakasuo, Michael},
  journal={International Journal of Social Robotics},
  volume={14},
  number={7},
  pages={1559--1581},
  year={2022},
  publisher={Springer}
}

@ARTICLE{howcroft2026,
    
AUTHOR={Howcroft, Alastair  and Giannaccini, Maria Elena  and Benford, Steve  and Khan, Ahmad  and Blake, Holly },
           
TITLE={Speech-touch integration for affective human–robot interaction: a scoping review},
          
JOURNAL={Frontiers in Robotics and AI},
          
VOLUME={13},
  
YEAR={2026},
  
URL={https://www.frontiersin.org/journals/robotics-and-ai/articles/10.3389/frobt.2026.1785039},
  
DOI={10.3389/frobt.2026.1785039},
  
ISSN={2296-9144},
}
}

\newpage

\vspace{11pt}

\vspace{11pt}

\vfill

\end{document}